\documentclass[conference]{IEEEtran}

\usepackage{amssymb,amsfonts,amsmath,amsthm,amscd,dsfont,mathrsfs}
\usepackage{graphicx,float,psfrag,epsfig}
\usepackage{wrapfig}
\usepackage{algorithm}
\usepackage{algorithmic}
\usepackage{rotating}
\usepackage{multirow}

\DeclareMathAlphabet{\mathpzc}{OT1}{pzc}{m}{it}

\def\eps{\epsilon}

\def\reals{{\mathds R}}

\def\tN{\widetilde{N}}

\def\diag{{\rm diag}}

\def\E{\mathbb E}
\def\R{\mathbb R}

\def\<{\langle}
\def\>{\rangle}
\def\diag{{\rm diag}}

\def\cP{{\cal P}}

\def\eps{\epsilon}

\def\hM{\widehat{M}}

\def\E{{\mathbb E}}

\def\rank{\rm{rank}}

\def\optspace{{\sc OptSpace }}
\def\incoptspace{{\sc Incremental OptSpace }}

\begin{document}

\title{Low-rank Matrix Completion with Noisy Observations: a Quantitative Comparison}

\author{
\IEEEauthorblockN{Raghunandan~H.~Keshavan, Andrea~Montanari${}^\dagger$ and Sewoong~Oh}
\IEEEauthorblockA{Electrical Engineering and $\dagger$Statistics Department \\
Stanford University, Stanford, CA 94304\\
\{raghuram,montanari,swoh\}@stanford.edu} 
}

\maketitle
%
%
\begin{abstract}
We consider a problem of significant practical importance, 
namely, the reconstruction of a low-rank data matrix
from a small subset of its entries.  
This problem appears in many areas such as 
collaborative filtering, computer vision and wireless sensor networks. 
In this paper, we focus on the matrix completion problem 
in the case when the observed samples are corrupted by noise.
We compare the performance of 
three state-of-the-art matrix completion algorithms 
(OptSpace, ADMiRA and FPCA) on a single simulation platform 
and present numerical results. We show that in practice 
these efficient algorithms can be used to
reconstruct real data matrices,
as well as randomly generated matrices, accurately.
\end{abstract}
%

%
%
\section{Introduction}
 We consider the problem of reconstructing an $m\times n$ 
low rank matrix $M$ from a small set of observed entries possibly corrupted by noise. 
This problem is of considerable practical interest and has many applications. 
One example is collaborative filtering, where users submit rankings 
for small subsets of, say, movies, and the goal is 
to infer the preference of unrated movies for a recommendation system \cite{Net06}.
It is believed that the movie-rating matrix is approximately low-rank, 
since only a few factors contribute to a user's preferences. 
Other examples of matrix completion include the problem of inferring 
3-dimensional structure from motion \cite{SFM} and
triangulation from incomplete 
data of distances between wireless sensors \cite{OKM09}.

%
%
\subsection{Prior and related work}\label{sec:relatedwork}

 On the theoretical side, most recent work focuses on 
algorithms for exactly recovering 
the unknown low-rank matrix and providing an upper bound on the number of 
observed entries that guarantee successful recovery with high probability, when the 
observed set is drawn uniformly at random over all subsets of the same size.
The main assumptions of this \emph{matrix completion problem with exact observations} is that 
the matrix $M$ to be recovered has rank $r\ll m,n$ and that the observed entries are known exactly.
Adopting techniques from compressed sensing, Cand{\`e}s and Recht introduced 
a convex relaxation to the NP-hard problem which is to find a minimum 
rank matrix matching the observed entries \cite{CaR08}. 
They introduced the concept of incoherence property and 
proved that for a matrix $M$ of rank $r$ which has the 
 incoherence property, solving the convex relaxation 
correctly recovers the unknown matrix, with high probability, 
if the number of observed entries $|E|$ satisfies, $|E| \geq Crn^{1.2}\log n$.

Recently \cite{KOM09} improved the bound to 
$|E| \geq C rn\max\{\log n, r\}$ with an extra condition that the matrix 
has bounded condition number, 
where the condition number of a matrix is defined as 
the ratio between the largest singular value and the smallest singular value of $M$. 
We introduced an efficient algorithm called {\sc OptSpace}, 
based on spectral methods 
followed by a local manifold optimization. 
For a bounded rank $r=O(1)$, the performance bound of {\sc OptSpace} 
 is order optimal \cite{KOM09}.
 Cand{\`e}s and Tao proved a similar bound $|E| \geq C nr(\log n)^6$
with a stronger assumption on the original matrix $M$,
known as the \emph{ strong incoherence condition} 
but without any assumption on the condition number of the matrix $M$ \cite{CandesTaoMatrix}.
For any value of $r$, it is only suboptimal by a poly-logarithmic factor.


 While most theoretical work focus on proving bounds for the
exact matrix completion problem, 
a more interesting and practical problem is 
when the matrix $M$ is only approximately low rank or 
when the observation is corrupted by noise. 
The main focus of this \emph{matrix completion with noisy observations}
 is to design an algorithm to find an $m\times n$ low-rank matrix $\hM$ 
that best approximates the original matrix $M$
and provide a bound on the root mean squared error (RMSE) given by,
\begin{eqnarray}
{\rm RMSE} =  \frac{1}{\sqrt{mn}}||M-\hM||_F\;.  \label{eq:RMSE}
\end{eqnarray}
Cand{\`e}s and Plan introduced a generalization of 
the convex relaxation from \cite{CaR08} to the noisy case, 
and provided a bound on the RMSE \cite{CandesPlan}. 
More recently, a bound on the RMSE 
achieved by the \optspace algorithm with noisy observations was obtained in \cite{KMO09noise}. 
This bound is order optimal in a number of situations 
and improves over the analogous result in \cite{CandesPlan}.
Detailed comparison of these two results are provided 
in Section \ref{sec:theoreticalcomparison}.

 On the practical side, directly solving the convex relaxation introduced in \cite{CaR08} 
requires solving a Semidefinite Program (SDP), the complexity of which grows proportional to $n^3$. 
Recently, many authors have proposed efficient algorithms 
for solving the low-rank matrix completion problem.
These include Accelerated Proximal Gradient (APG) algorithm \cite{APG},
Fixed Point Continuation with Approximate SVD (FPCA) \cite{FPCA},
Atomic Decomposition for Minimum Rank Approximation ({\sc ADMiRA}) \cite{ADMiRA}, 
{\sc Soft-Impute} \cite{MHT09}, 
Subspace Evolution and
Transfer (SET) \cite{DM09}, 
Singular Value Projection (SVP) \cite{MJD09}, and {\sc OptSpace} \cite{KOM09}.
 In this paper, we provide numerical comparisons of the performance of 
three state-of-the-art algorithms, namely, {\sc OptSpace}, {\sc ADMiRA} and  FPCA,
and show that these efficient algorithms can be used to
reconstruct real data matrices,
as well as randomly generated matrices, accurately.

%
%
\subsection{Outline}\label{sec:outline}

 The organization of this paper is as follows. In Section 2,
we describe the matrix completion problem and 
efficient algorithms to solve the matrix completion problem 
when the observations are corrupted by noise. 
Section 3 discusses the results of numerical simulations 
and compares the performance of 
three matrix completion algorithms with 
respect to speed and accuracy. 

%
%
\section{The model definition and algorithms} \label{sec:model}


\subsection{Model definition} \label{sec:modeldef}
The matrix $M$ has dimensions $m \times n$, and we define $\alpha\equiv m/n$ to denote the ratio. 
In the following we assume, without loss of generality, $\alpha\geq 1$.
We assume that the matrix $M$ has exact low rank $r\ll n$, 
that is, there exist matrices $U$ of dimensions $m\times r$, 
$V$ of dimensions $n \times r$, and a diagonal matrix $\Sigma$ 
of dimensions $r\times r$, such that 
\begin{eqnarray*}
 M = U\Sigma V^T\; .
\end{eqnarray*}
Notice that for a given matrix $M$, the factors $(U,V,\Sigma)$ are not unique.
Further, each entry of $M$ is perturbed, thus producing an 
`approximately' low-rank matrix $N$, with 
\begin{eqnarray*}
 N_{ij} = M_{ij}+Z_{ij}\; ,
\end{eqnarray*}
where the matrix $Z$ accounts for the noise.

Out of the $m\times n$ entries of $N$, a subset $E\subseteq[m]\times[n]$ is observed.
Let $N^E$ be the $m \times n$ observed matrix with all the observed values, such that
\begin{eqnarray*}
  N^E_{ij} = \left\{
              \begin{array}{rl}
              N_{ij} & \text{if } (i,j)\in E\, ,\\
              0       & \text{otherwise.}
              \end{array} \right. 
\end{eqnarray*}
Our goal is to find a low rank estimation $\hM(N^E,E)$ of the original matrix $M$ from
the observed noisy matrix $N^E$ and the set of observed indices $E$.
%
%
\subsection{Algorithms} \label{sec:algorithms}

In the case when there is no noise, that is $N_{ij}=M_{ij}$, 
solving the following optimization problem will recover the original matrix correctly,
if the number of observed entries $|E|$ is large enough.
\begin{eqnarray}
&\text{minimize}   & \rank(X)       \; \label{P0} \\
&\text{subject to} & \cP_E(X)=\cP_E(M)      \;, \nonumber
\end{eqnarray}
where $X\in\reals^{m\times n}$ is the variable matrix, $\rank(X)$ is the rank of matrix $X$, 
and $\cP_E(\cdot)$ is the projector operator defined as 
\begin{eqnarray}
\cP_E(M)_{ij} = \left\{\begin{array}{ll}
M_{ij} & \mbox{ if $(i,j)\in E$,}\\
0 & \mbox{otherwise.}
\end{array}\right.\label{eq:ProjectorDef}
\end{eqnarray}
This problem finds the matrix with the minimum rank that matches all the observations. 
Notice that the solution of problem (\ref{P0}) is optimal. If this problem does not recover the correct matrix $M$ then there exists at least one other rank-$r$ matrix that matches all the observations 
and no other algorithm can distinguish which one is the correct solution. 
However, this optimization problem is NP-hard and
all known algorithms require doubly exponential time in $n$ \cite{CaR08}.

In compressed sensing, minimizing the $l_1$ norm of a vector is the tightest convex relaxation of 
minimizing the $l_0$ norm, or equivalently minimizing the number of non-zero entries, 
for sparse signal recovery. We can adopt this idea to matrix completion, where
$\rank(\cdot)$ of a matrix corresponds to $l_0$ norm of a vector, 
and nuclear norm to $l_1$ norm \cite{CaR08}, 
where the nuclear norm of a matrix is defined as the sum of its singular values.
\begin{eqnarray}
&\text{minimize}   & ||X||_*       \; \label{P1}\\
&\text{subject to} & \cP_E(X)=\cP_E(M)      \;, \nonumber
\end{eqnarray}
where $||X||_*$ denotes the nuclear norm of $X$.

In this paper, we are interested in the more practical case when 
the observations are contaminated by noise or the original matrix 
to be reconstructed is only approximately low rank.
In this case, the constraint $\cP_E(X)=\cP_E(M)$ 
must be relaxed. This results in either the problem \cite{CandesPlan,FPCA,APG,MHT09}
\begin{eqnarray}
&\text{minimize}   & ||X||_*       \; \label{P2}\\
&\text{subject to} & ||\cP_E(X)-\cP_E(N)||_F \leq \Theta      \;, \nonumber
\end{eqnarray}
or its Lagrangian version
\begin{eqnarray}
&\text{minimize}   & \mu||X||_*+\frac{1}{2}||\cP_E(X)-\cP_E(N)||_F^2    \;. \label{P3}
\end{eqnarray}
In the following, we briefly explain the objective of the three state-of-the-art 
matrix completion algorithms basaed on the relaxation, namely, 
FPCA, {\sc ADMiRA}, and {\sc OptSpace}. 

FPCA, introduced in \cite{FPCA}, 
is an efficient algorithms for solving the convex relaxation, 
which is a nuclear norm regularized least squares problem in (\ref{P3}).
Following the same line of argument given in \cite{CandesPlan}, 
we choose $\mu=\sqrt{2np}\sigma$, where $p=|E|/mn$ and $\sigma^2$
is the variance of each entry in $Z$.

{\sc ADMiRA}, introduced in  \cite{ADMiRA}, 
is an efficient algorithm which is based on 
the atomic decomposition and extends the idea of the 
Compressive Sampling Matching Pursuit (CoSaMP) \cite{CoSaMP}. 
{\sc ADMiRA}  is an iterative method for solving the following 
rank-$r$ matrix approximation problem.
\begin{eqnarray}
&\text{minimize}   & ||\cP_E(X)-\cP_E(N)||_F  \; \label{P4}\\
&\text{subject to} & \rank(X)\leq r     \;. \nonumber
\end{eqnarray}
One drawback of {\sc ADMiRA} is that 
it requires the prior knowledge of the rank
of the original matrix $M$.
In the following numerical simulations, for fair comparison, 
we first run a rank estimation algorithm 
to guess the rank of the original matrix and 
use the estimated rank in {\sc ADMiRA}. 
The rank estimation algorithm is explained in the next section.

{\sc OptSpace}, introduced in \cite{KOM09}, 
is a novel and  efficient algorithm based on the spectral 
method followed by a local optimization,
 which consists of the following three steps.\\
 1. Trim the matrix $N^E$.\\
 2. Compute the rank-$r$ projection of the trimmed observation matrix.\\
 3. Minimize $||\cP_E(XSY^T)-\cP_E(N)||_F^2$ through gradient descent, using the rank-$r$ projection as the initial guess.

In the trimming step, we set to zero
all columns in $N^E$ with the number of samples larger than 
$2|E|/n$ and set to zero all rows with the number of samples larger than 
$2|E|/m$.
In the second step, the rank-$r$ projection of a matrix $A$ is defined as 
\begin{eqnarray}
	\cP_r(A)=\frac{mn}{|E|}\sum_{i=1}^r \sigma_i x_i y_i^T \;, \label{eq:rankrprojection}
\end{eqnarray}
where the SVD of $A$ is given by 
$A= \sum_{i=1}^n \sigma_i x_i y_i^T$.
The basic idea is that the rank-$r$ projection of the trimmed observation matrix 
provides an excellent initial guess,
so that the standard gradient descent provides 
a good estimate after this initialization.
Note that  we need to estimate the target rank $r$.
To estimate the target rank $r$ for {\sc ADMiRA} and {\sc OptSpace}, 
we used the following simple rank estimation procedure.

%
%
\subsection{Rank estimation algorithm}\label{sec:rankestimation}

Let $\tN^E$ be the trimmed version of $N^E$.
By singular value decomposition of the trimmed matrix, we have
\begin{equation*}
\tN^E = \sum_{i=1}^{\min(m,n)}\sigma_i  x_iy_i^T\, , 
\end{equation*}
where $x_i$ and $y_i$ are the left and right singular vectors 
corresponding to $i$th singular value $\sigma_i$.
Then, the following cost function is defined in terms of the singular values.
\begin{eqnarray*}
 R(i)=\frac{\sigma_{i+1}+\sigma_1\sqrt{\frac{i\sqrt{mn}}{|E|}}}{\sigma_i} \;.
\end{eqnarray*}
Finally, the estimated rank is the index $i$ that minimizes the cost function $R(i)$.

The idea behind this algorithm is that if enough entries of $N$ are revealed 
and there is little noise 
then there is a clear separation between the first $r$ singular values, 
which reveal the structure of the matrix $M$ to be reconstructed, 
and the spurious ones \cite{KOM09}. Hence, $\sigma_{i+1}/\sigma_i$ is
minimum when $i$ is the correct rank $r$. 
The second term is added to ensure the robustness of the algorithm.

%
%
\subsection{Comparison of the performance guarantees}\label{sec:theoreticalcomparison}

 Performance guarantees for matrix completion problem 
 with noisy observations are proved in \cite{CandesPlan} and \cite{KMO09noise}. 
 Theorem 7 of \cite{CandesPlan} shows the following bound on the performance of solving convex relaxation (\ref{P2}) under some constraints on the matrix $M$
 known as the strong incoherence property.
 \begin{eqnarray}
 	{\rm RMSE} \leq 7\sqrt{\frac{n}{|E|}}||\cP_E(Z)||_F + \frac{2}{n\sqrt{\alpha}} ||\cP_E(Z)||_F \;,
	\label{eq:bound1}
 \end{eqnarray} 
 where RMSE is defined in Eq.~(\ref{eq:RMSE}).
The constant in front of the first term is in fact slightly smaller than $7$ 
in \cite{CandesPlan},
but in any case larger than $4\sqrt{2}$.

Theorem 1.2 of \cite{KMO09noise} shows the following bound on the performance of {\sc OptSpace} under the assumptions that $M$ is incoherent and has a bounded 
condition number $\kappa= \sigma_1(M)/\sigma_r(M)$, 
where the condition number of a matrix is defined as 
the ratio between the largest singular value $\sigma_1(M)$ and the smallest singular value $\sigma_r(M)$ of $M$.
 \begin{eqnarray}
 	{\rm RMSE} \leq  C\kappa^2\sqrt{\alpha r}\frac{n}{|E|} ||\cP_E(Z)||_2\;,
	\label{eq:bound2}
 \end{eqnarray} 
for some numerical constant $C$.

Although the assumptions on the above two theorems are not directly comparable, 
as far as the error bounds are concerned,  the bound (\ref{eq:bound2}) improves
over the bound (\ref{eq:bound1}) in several respects: 
 The bound (\ref{eq:bound2}) does not have the second term in the bound (\ref{eq:bound1}) which actually grows with the number of observed entries;
 The bound (\ref{eq:bound2}) decreases as $n/|E|$ rather than $(n/|E|)^{1/2}$;
 The bound (\ref{eq:bound2}) is proportional to the operator norm of the noise matrix 
$||\cP_E(Z)||_2$ instead of the Frobenius norm $||\cP_E(Z)||_F \geq ||\cP_E(Z)||_2$.
For $E$ uniformly random, one expects $||\cP_E(Z)||_F$ to be roughly of order $||\cP_E(Z)||_2\sqrt{n}$. For instance, if the entries of $Z$ are i.i.d. Gaussian with bounded variance $\sigma$, $||\cP_E(Z)||_F=\Theta(\sqrt{|E|})$ while $||\cP_E(Z)||_2$ is of 
order $\sqrt{|E|/n}$.

In the following, we 
 numerically compare the performances  of 
 three efficient algorithms, {\sc OptSpace}, {\sc ADMiRA} and FPCA, for solving the matrix completion problem,
with real data matrices as well as randomly generated matrices.

%
%
\section{Numerical results}\label{sec:Implementation}

 In this section, we present numerical comparisons between three
 approximate low-rank matrix completion algorithms : 
 {\sc OptSpace}, {\sc ADMiRA} and FPCA.
 The performance of each algorithm is compared in terms of 
 the relative root mean squared error defined as in Eq.~({\ref{eq:RMSE}}).
 for randomly generated matrices in Section \ref{sec:synthetic} 
 and real data matrices in Section \ref{sec:real}.
 We used MATLAB implementations of the algorithms
 and tested them on a 3.0 GHz Desktop computer with 2 GB RAM.
 FPCA is available from www.columbia.edu/$\sim$sm2756/FPCA.htm
 and {\sc OptSpace}  is available from www.stanford.edu/$\sim$raghuram/optspace/ .
 
%
%
\subsection{Numerical results with randomly generated matrices}
\label{sec:synthetic}

For numerical simulations with randomly generated matrices, 
we use $n\times n$ test matrices $M$ of rank $r$ generated as 
$M = UV^T$, where $U$ and $V$ are $n\times r$ matrices with 
each entry being sampled independently from a standard 
Gaussian distribution ${\cal N}(0,1)$, unless specified otherwise.
Each entry is revealed independently with probability $\eps/n$, so that 
on an average $n\eps$ entries are revealed. 
The observation is corrupted by added noise matrix $Z$, so that the observation
for the index $(i,j)$ is $M_{ij}+Z_{ij}$.

In the standard scenario, 
we typically make the following three assumptions on the noise matrix $Z$.
(1) The noise $Z_{ij}$ does not depend on the value of the matrix $M_{ij}$.
(2) The entries of $Z$, $\{Z_{ij}\}$, are independent.
(3) The distribution of each entries of $Z$ is Gaussian.
The above matrix completion algorithms are expected to be especially effective 
under this standard scenario for the following two reasons. 
First, the squared error objective function that the algorithms minimize is 
well suited for the Gaussian noise. Second, the independence of $Z_{ij}$'s 
ensure that the noise matrix is almost full rank and 
the singular values are evenly distributed. This implies that 
for a given noise power $||Z||_F$, 
the spectral norm $||Z||_2$ is much smaller than $||Z||_F$.
In the following, we fix $m=n=500$ and $r=4$, 
and study how the performance changes with different noise.
Each of the simulation results is averaged over 10 instances
and is shown with respect to two basic parameters,
the average number of revealed entries per row $\eps$ and 
the signal-to-noise ratio, ${\rm SNR}= \E[||M||^2_F]/\E[||Z||^2_F] $.

%
%
\subsubsection{Standard scenario}

\begin{figure}
\begin{center}
\includegraphics[width=8cm]{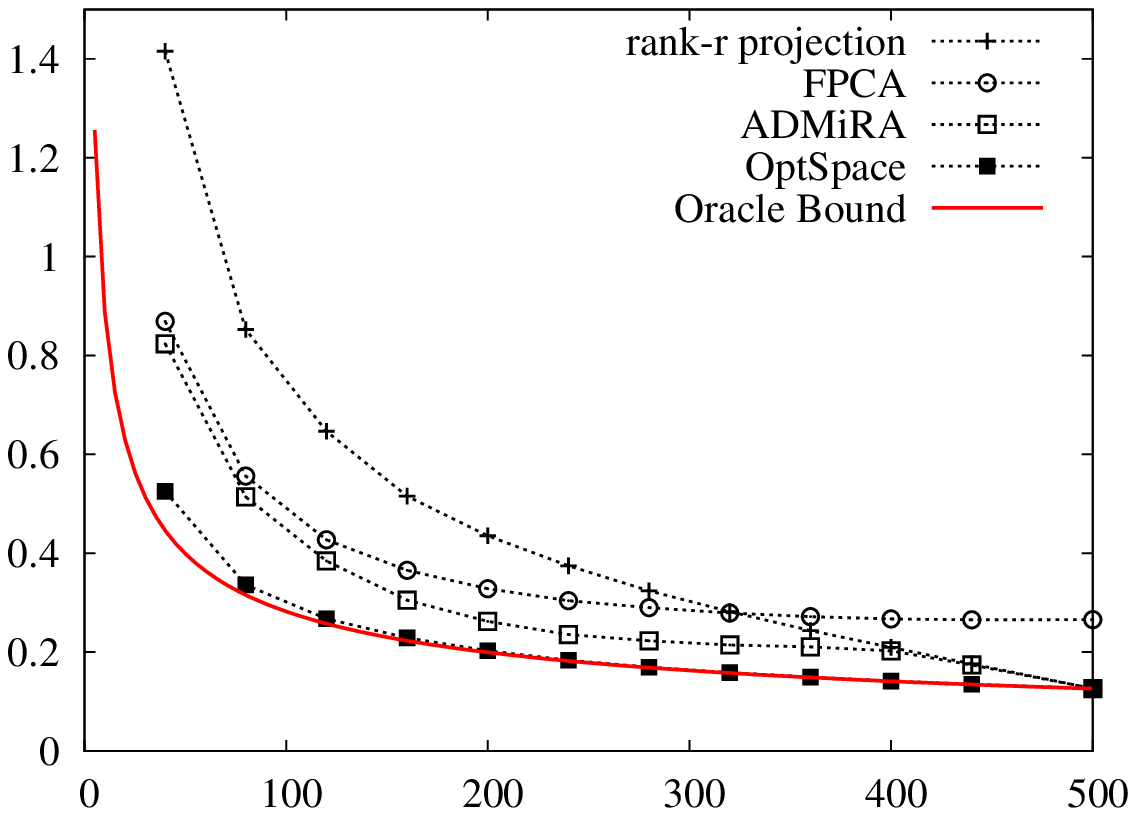}
\put(-106,0){\scriptsize{$\eps$}}
\put(-220,80){\begin{sideways}\scriptsize{RMSE}\end{sideways}}\\
\hspace{-0.3cm}
\includegraphics[width=8.3cm]{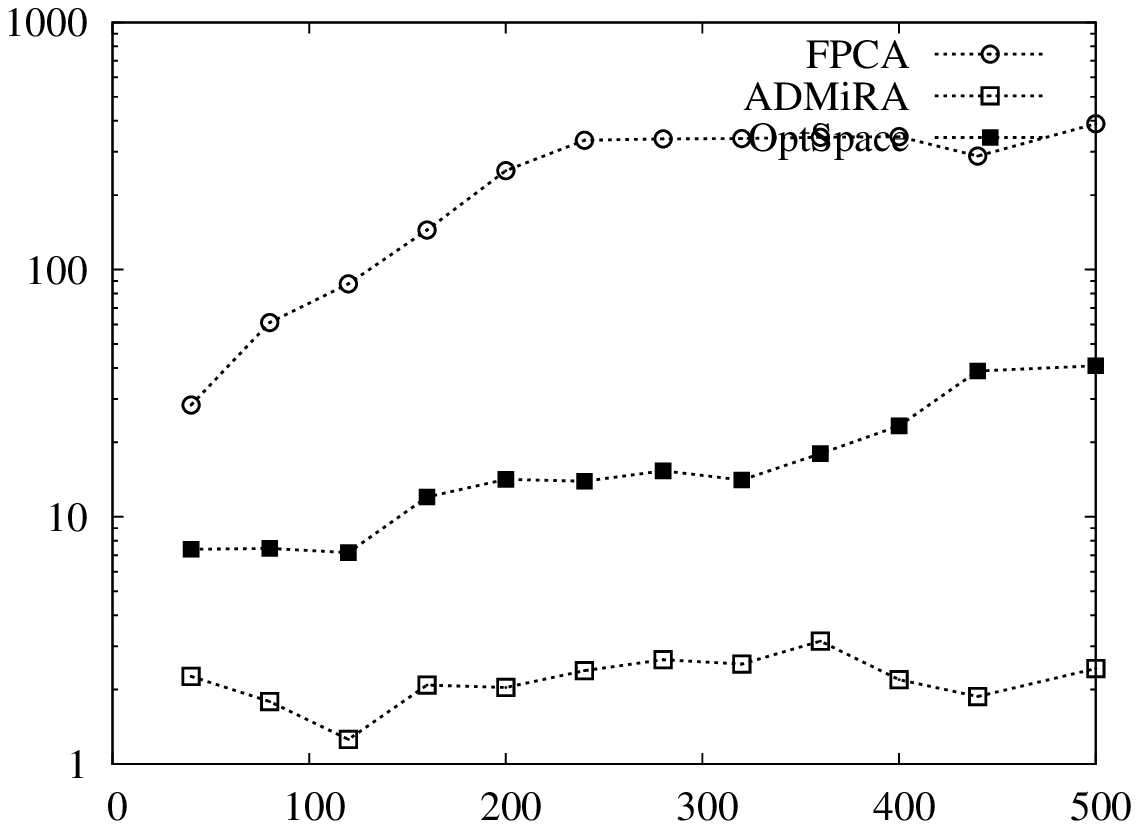}
\put(-106,0){\scriptsize{$\eps$}}
\put(-220,74){\begin{sideways}\scriptsize{seconds}\end{sideways}}\\
\caption{{\small The RMSE (above) and  the computation time in seconds (below) as a function of the average number of observed entries per row $\eps$ for SNR=$4$ under the standard scenario. }} \label{fig:noise0_eps}
\end{center}
\end{figure}

In this standard scenario, the noise $Z_{ij}$'s are distributed as 
i.i.d. Gaussian {\cal N}(0,$\sigma^2$). 
Note that the SNR is equal to $4/\sigma^2$.
There is a basic trade-off between two metrics of interest:
the accuracy of the estimation is measured using RMSE and 
 the computation complexity is measured by the running time in seconds.

In order to interpret the simulation results, 
they are compared to the RMSE achieved by the oracle
and a simple rank-$r$ projection algorithm defined as Eq.~(\ref{eq:rankrprojection}).
The rank-$r$ projection algorithm simply computes $\cP_r(N^E)$.
The oracle has prior knowledge of  
the linear subspace spanned by $\{UX^T+YV^T:X\in\R^{m\times r},Y\in\R^{n\times r}\}$, and 
the RMSE of the oracle estimate is $\sigma\sqrt{(2nr-r^2)/n\eps}$  \cite{CandesPlan}.

Figure \ref{fig:noise0_eps} shows the performance and the computation time for each of 
the algorithms with respect to $\eps$ under the standard scenario for fixed SNR$=4$. 
For most values of $\eps$, the simple rank-$r$ projection has the worst performance. 
However, when all the entries are revealed and the noise is i.i.d. Gaussian, 
the rank-$r$ projection coincides with the oracle bound, 
which in this simulation corresponds to the value $\eps=500$. 
Note that the behavior of the performance curves of FPCA, {\sc ADMiRA}, 
and {\sc OptSpace} with respect to $\eps$ is similar to the oracle bound, 
which is proportional to $1/\sqrt{\eps}$. 

Among the three algorithms, FPCA has the largest RMSE, 
and {\sc OptSpace} is very close to the oracle bound for all values of $\eps$. 
Note that when all the values are revealed, {\sc ADMiRA} is an efficient way of implementing rank-$r$ projection, and the performances are expected to be similar. 
This is confirmed by the observation that for $\eps \geq 400$ the two curves are almost identical. One of the reasons why the RMSE of FPCA 
does not decrease with $\eps$ for large values of $\eps$ is that 
FPCA overestimates the rank and returns estimated matrices 
with rank much higher than $r$, whereas the rank estimation algorithm
 used for {\sc ADMiRA} and {\sc OptSpace} always returned the 
 correct rank $r$ for $\eps \geq 80$. 

The second figure in Figure \ref{fig:noise0_eps} shows 
the average running time of the algorithms with respect to $\eps$.
Note that due to the large difference between the running time of three algorithms, 
the time is displayed in log scale.
For most of the simulations, {\sc ADMiRA} had shortest running time and {FPCA} the longest, and the gap was noticeably large as clearly shown in the figure.
For FPCA and {\sc OptSpace}, the computation time increased with $\eps$, whereas 
{\sc ADMiRA} had relatively stable computation time independent of $\eps$.

\begin{figure}
\begin{center}
\includegraphics[width=8cm]{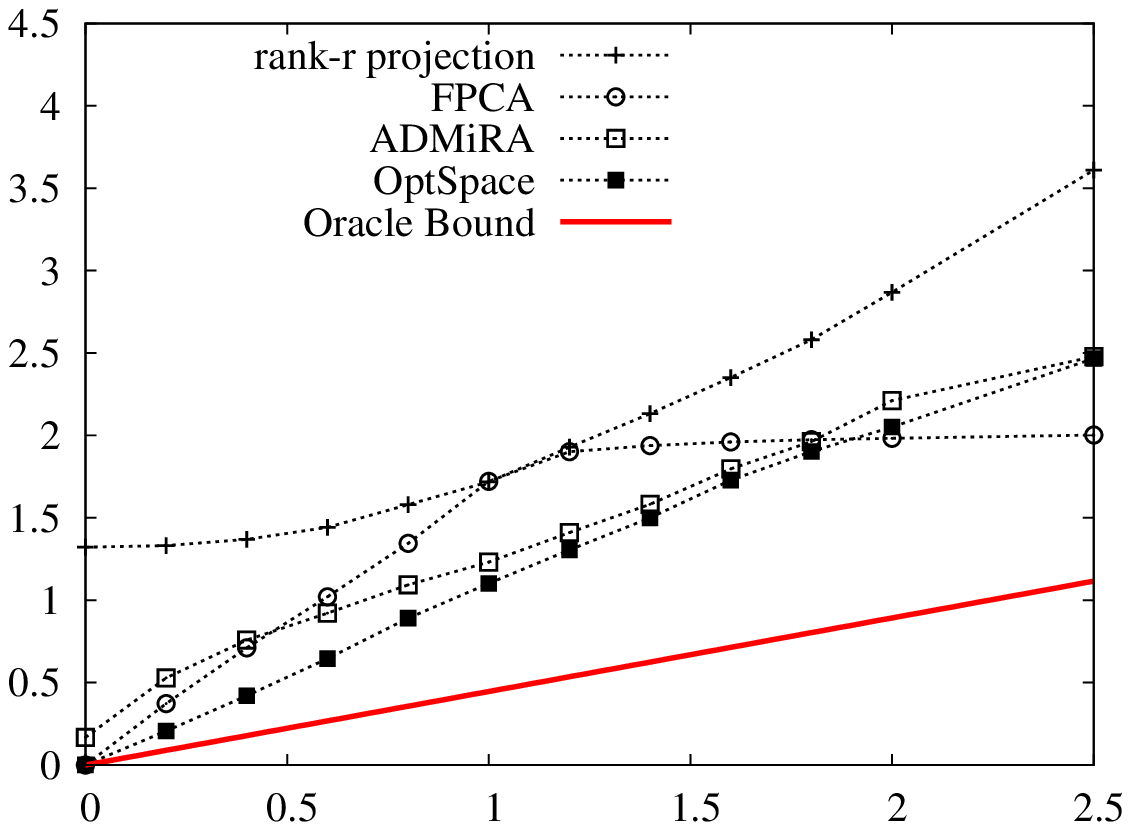}
\put(-121,0){\scriptsize{$1/\sqrt{\rm SNR}$}}
\put(-220,80){\begin{sideways}\scriptsize{RMSE}\end{sideways}}\\
\hspace{-0.1cm}
\includegraphics[width=8.1cm]{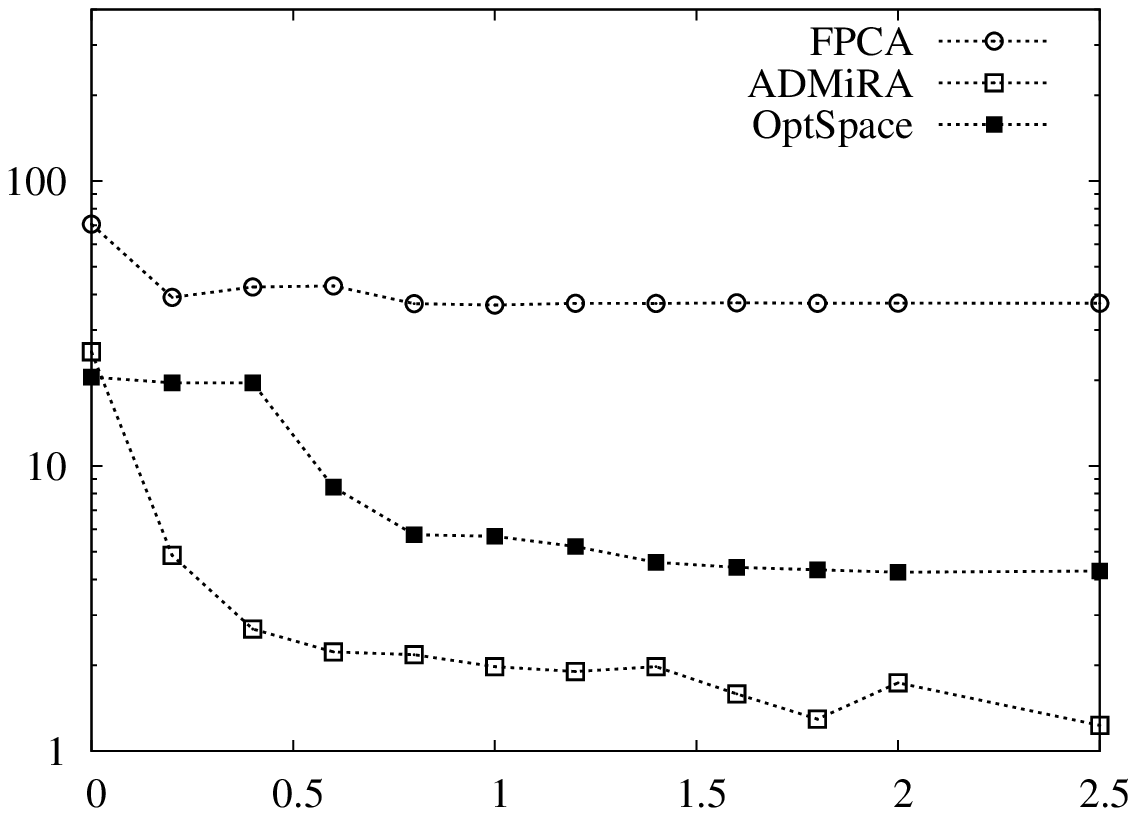}
\put(-121,0){\scriptsize{$1/\sqrt{\rm SNR}$}}
\put(-220,74){\begin{sideways}\scriptsize{seconds}\end{sideways}}\\
\caption{{\small The RMSE (above) and the computation time in seconds (below) as a function of $1/{\rm SNR}$ for fixed $\eps=40$ under the standard scenario.}} \label{fig:noise0_sig}
\end{center}
\end{figure}

Figure \ref{fig:noise0_sig} show the performance and computation time for each of 
the algorithms against the SNR 
within the standard scenario for fixed $\eps=40$. 
The behavior of the performance curves of {\sc ADMiRA} and {\sc OptSpace} are
similar to the oracle bound which is linear in $\sigma$ 
which, in the standard scenario, is equal to $2/\sqrt{{\rm SNR}}$. 
The performance of the rank-$r$ projection algorithm is 
determined by two factors. One is the added noise which is linear in $\sigma$
and  the other is the error caused by the erased entries which is 
constant independent of SNR. These two factors add up, whence the 
performance curve of the rank-$r$ projection follows. 
The reason the RMSE of FPCA does not decrease with SNR
for values of SNR less than $1$ is not that the estimates are good but
rather the estimated entries gets very small and the resulting RMSE is close to 
$\sqrt{\E[||M||_F^2/n^2]}$, which is $2$ in this simulation, regardless of the noise power.
When there is no noise, which corresponds to the value $1/{\rm SNR}=0$, 
FPCA and {\sc OptSpace} both recover the original matrix correctly 
for this chosen value of $\eps=40$.
For all three algorithms, the computation time is larger for smaller noise,
and the reason is that it takes more iterations until the stopping criterion is met.
Also, for most of the simulations with different SNR, 
{\sc ADMiRA} had shortest running time and 
{FPCA} the longest.

%
%
\subsubsection{Multiplicative Gaussian noise}

\begin{figure}
\begin{center}
\includegraphics[width=8cm]{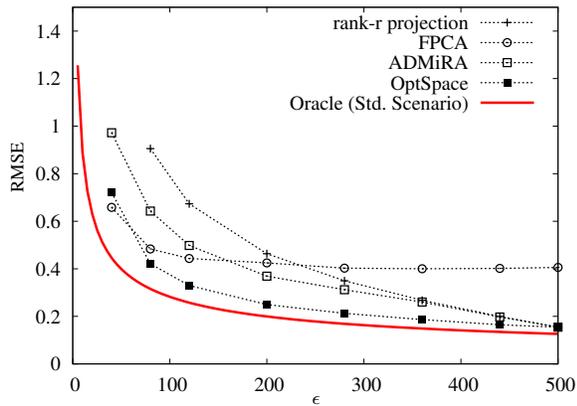}
\put(-106,0){\scriptsize{$\eps$}}
\put(-220,80){\begin{sideways}\scriptsize{RMSE}\end{sideways}}
\hspace{-0.3cm}
\caption{{\small The RMSE as a function of the average number of observed entries per row $\eps$ for fixed SNR$=4$ within the multiplicative noise model. }} \label{fig:noise_multi}
\end{center}
\end{figure}

In sensor network localization \cite{WZBY06}, where the entries of the matrix corresponds to 
the pair-wise distances between the sensors, the observation noise is oftentimes 
assumed to be multiplicative. In formulae, $Z_{ij}=\xi_{ij}M_{ij}$, 
where $\xi_{ij}$'s are distributed as i.i.d. Gaussian with zero mean.
The variance of $\xi_{ij}$'s are chosen to be $1/r$ 
so that the resulting noise power is one.
Note that in this case, $Z_{ij}$'s are mutually dependent through $M_{ij}$'s and 
the values of the noise also depend on the value of the matrix entry $M_{ij}$.

Figure \ref{fig:noise_multi} shows the RMSE with respect to $\eps$ under multiplicative Gaussian noise. The RMSE of the rank-$r$ projection for $\eps=40$
is larger than $1.5$ and is omitted in the figure.
The bottommost line corresponds to the oracle performance under standard scenario,
and is displayed here, and all of the following figures, to serve as a reference for comparison.
The main difference with respect to Figure \ref{fig:noise0_eps} is that 
all the performance curves are larger under multiplicative noise.
For the same value of SNR, 
it is more difficult to distinguish the noise from the original matrix,
since the noise is now correlated with the matrix $M$.

%
%
\subsubsection{Outliers}

\begin{figure}
\begin{center}
\includegraphics[width=8cm]{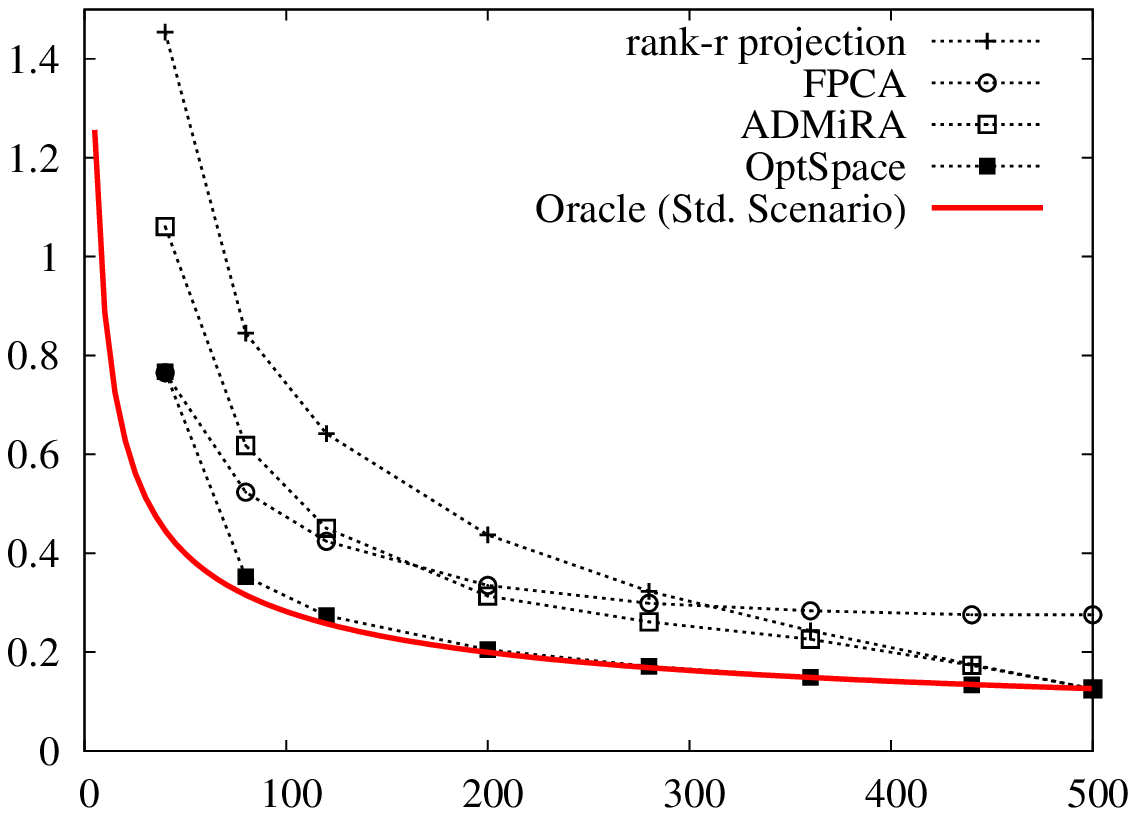}
\put(-106,0){\scriptsize{$\eps$}}
\put(-220,80){\begin{sideways}\scriptsize{RMSE}\end{sideways}}\\
\includegraphics[width=8cm]{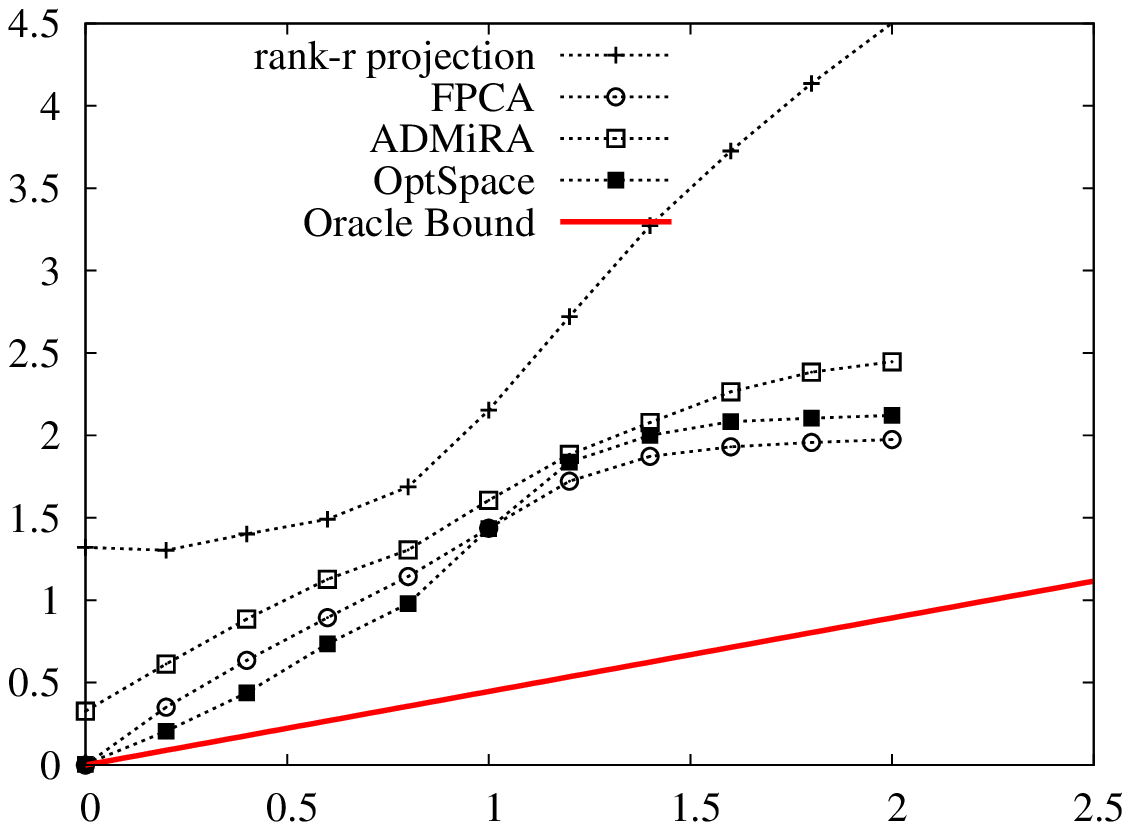}
\put(-121,0){\scriptsize{$1/\sqrt{\rm SNR}$}}
\put(-220,80){\begin{sideways}\scriptsize{RMSE}\end{sideways}}
\hspace{-0.3cm}
\caption{{\small The RMSE as a function of the average number of observed entries per row $\eps$ for fixed SNR$=4$ with outliers (above) and the RMSE as a function of $1/{\rm SNR}$ for fixed $\eps=40$ with outliers (below).}} \label{fig:noise_outlier}
\end{center}
\end{figure}

In structure from motion \cite{SFM}, the entries of the matrix corresponds to 
the position of points of interest in $2$-dimensional images captured by 
cameras in different angles and locations. 
However, due to failures in the feature extraction algorithm, 
some of the observed positions are corrupted by large noise 
where as most of the observations are noise free.
To account for such outliers, we use the following model. 
\begin{eqnarray*}
  Z_{ij} = \left\{
              \begin{array}{rl}
              a  & \text{with probability $1/200$ } \, ,\\
              -a & \text{w.p. $1/200$ } \, ,\\
              0       & \text{w.p. $99/100$}\, .
              \end{array} \right. 
\end{eqnarray*}
The value of $a$ is chosen according to the target SNR$=400/a^2$.
This is clearly independent of the matrix entries and 
$Z_{ij}$'s are mutually independent, but the distribution is now non-Gaussian. 

Figure \ref{fig:noise_outlier} shows the performance of 
the algorithms with respect to $\eps$ and the SNR with outliers.
Comparing the first figure to Figure \ref{fig:noise0_eps}, we can see that
the performance for large value of $\eps$ is less affected by outliers compared to 
the small values of $\eps$. The second figure clearly shows how the performance degrades for non-Gaussian noise when the number of samples is small.  
The algorithms minimize the squared error $||\cP_E(X)-\cP_E(N)||_F^2$ as in (\ref{P3}) and (\ref{P4}).
For outliers, a suitable algorithm would be to 
minimize the $l_1$-norm of the errors instead of the $l_2$-norm. 
Hence, for this simulation outliers, we can see that the performance of the rank-$r$ 
projection, {\sc ADMiRA} and {\sc OptSpace} is worse than the Gaussian noise case.
However,  the performance of FPCA is almost the same as in the standard scenario. 

%
%
\subsubsection{Quantization noise}

\begin{figure}
\begin{center}
\includegraphics[width=8cm]{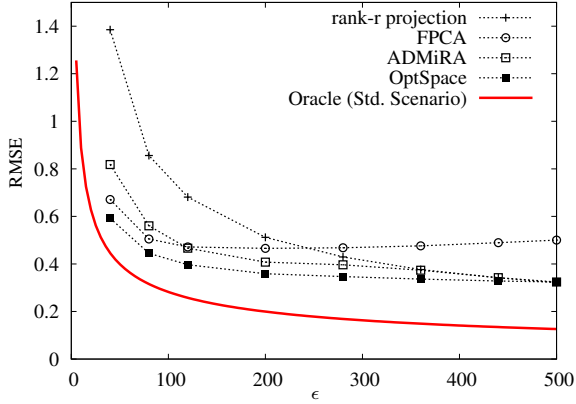}
\put(-106,0){\scriptsize{$\eps$}}
\put(-220,80){\begin{sideways}\scriptsize{RMSE}\end{sideways}}\\
\caption{{\small The RMSE as a function of the average number of observed entries per row $\eps$ for fixed SNR$=4$ with quantization.}} \label{fig:noise_quantization}
\end{center}
\end{figure}

One common model for noise is the quantization noise. 
For a regular quantization, we choose a parameter $a$ and 
quantize the matrix entries to the nearest value in  $\{\ldots, -a/2,a/2,3a/2,5a/2 \ldots\}$.
The parameter $a$ is chosen carefully such that the resulting SNR is $4$.
The performance for this quantization is expected to be worse than the multiplicative noise case, since now the noise is deterministic and completely depends on the matrix entries $M_{ij}$, whereas in the multiplicative noise model it was random.

Figure \ref{fig:noise_quantization} shows the performance against $\eps$
within quantization noise. The overall behavior of the performance curves is similar to Figure
\ref{fig:noise0_eps}, but all the curves are shifted up. Note that the bottommost line 
is the oracle performance in the standard scenario which is the same in all the figures.
Compared to Figure \ref{fig:noise_multi}, for the same values of SNR$=4$,  
quantization is much more damaging than the multiplicative noise as expected.

%
%
\subsubsection{Ill conditioned matrices}

\begin{figure}
\begin{center}
\includegraphics[width=8cm]{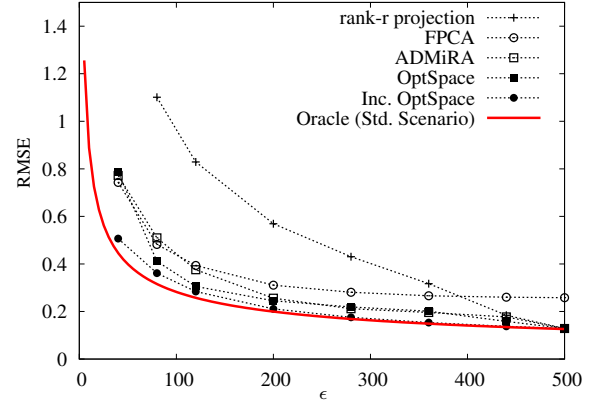}
\put(-106,0){\scriptsize{$\eps$}}
\put(-220,80){\begin{sideways}\scriptsize{RMSE}\end{sideways}}\\
\caption{{\small The RMSE as a function of the average number of observed entries per row $\eps$ for fixed SNR$=4$ with ill-conditioned matrices.}} \label{fig:noise_ill}
\end{center}
\end{figure}

In this simulation, we look at how the performance degrades under the standard scenario 
if the matrix $M$ is ill-conditioned. $M$ is generated as 
$M=\sqrt{4/166}\, U \,\diag([1,4,7,10])\, V^T$, where $U$ and $V$ are generated 
as in the standard scenario. The resulting matrix has condition number $10$
and the normalization constant $\sqrt{4/166}$ is chosen such that $\E[||M||_F]$ 
is the same as in the standard case. 

Figure \ref{fig:noise_ill} shows the performance with respect to $\eps$
with ill-conditioned matrix $M$. 
The performance of {\sc OptSpace} is similar to that of {\sc ADMiRA} for many values of $\eps$. However, a modification of {\sc OptSpace} called {\sc Incremental OptSpace} 
achieves a better performance in this case of ill-conditioned matrix.
The {\sc Incremental OptSpace} algorithm starts from finding a rank-$1$ approximation
from $N^E$ and incrementally finds higher rank approximations and 
has more robust performance when $M$ is ill-conditioned, but is computationally more expensive.

%
%

\subsection{Numerical results with real data matrices}
\label{sec:real}

In this section, we consider the low-rank matrix completion problems in the context of recommender systems, 
based on two real data sets : 
the Jester joke data set \cite{jesterjokes} and the Movielens data set \cite{movielens}. 
The Jester joke data set contains $4.1\times10^6$ ratings for 
100 jokes from 73,421 users. \footnote{The dataset is available at {http://www.ieor.berkeley.edu/$\sim$goldberg/jester-data/}} Since the number of users 
is large compared to the number of jokes, we randomly select 
$n_u\in\{100,1000,2000,4000\}$ users for comparison purposes.
As in \cite{FPCA}, we randomly choose two ratings for each user as a test set, 
and this test set, which we denote by $T$, 
is used in computing the prediction error in Normalized Mean Absolute Error (NMAE). 
The Mean Absolute Error (MAE) is defined as in \cite{FPCA,eigentaste}. 
\begin{eqnarray*}
 MAE = \frac{1}{|T|}\sum_{(i,j)\in T} |M_{ij}-\hM_{ij}| \;,
\end{eqnarray*}
where $M_{ij}$ is the original rating in the data set and $\hM_{ij}$ 
is the predicted rating for user $i$ and item $j$.
The Normalized Mean Absolute Error (NMAE) is defined as 
\begin{eqnarray*}
 NMAE = \frac{MAE}{M_{\rm max}-M_{\rm min}} \;,
\end{eqnarray*}
where $M_{\rm max}$ and $M_{\rm min}$ are upper and lower bounds for the ratings. 
In the case of Jester joke, all the ratings are in $[-10,10]$ which implies that 
$M_{\rm max}=10$ and $M_{\rm min}=-10$. 

\vspace{0.4cm}
\begin{tabular}{ c c | c | c | c }
\hline
$n_u$ & $n_s$  & {\sc OptSpace} & {FPCA} & {{\sc ADMiRA}}\\
\hline
$100$ & 
 $7484$ & 
$0.17674 $ & 
$0.20386 $ & 
$0.18194 $ \\

$1000$ & 
$73626$ & 
$0.15835 $ & 
$0.16114 $ & 
$0.16194 $ \\

$2000$ & 
$146700$ & 
$0.15747 $ & 
$0.16101 $ & 
$0.16286 $ \\

$4000$ & 
$290473$ & 
$0.15918 $ &
$0.16291 $ &
$0.16317 $ \\
\hline
$943$ & 
$80000$ & 
$0.18638 $ &
$0.19018 $ &
$0.24276 $  \\
\hline

\end{tabular}
\vspace{0.3cm}

The numerical results for Jester joke data set using {\sc Incremental OptSpace}, 
{\sc FPCA} and {\sc ADMiRA} are presented in the first four columns of the table above.
The number of jokes $m$ is fixed at 100 and the number of users $n_u$ and the number of samples $n_s$ is given in the first two columns. The resulting NMAE of each algorithm is
shown in the table. 
To get an idea of how good the predictions are, 
consider the case where each missing entry is predicted with a random number 
drawn uniformly at random in $[-10,10]$ and the actual rating is also 
a random number with same distribution.
After a simple computation, we can see that 
the resulting NMAE of the random prediction is 0.333.
As another comparison, for the same data set with $n_u=18000$, 
simple nearest neighbor algorithm and {\sc Eigentaste} both 
yield NMAE of 0.187 \cite{eigentaste}. The NMAE of \incoptspace is lower than 
these simple algorithms even for $n_u=100$ and tends to decrease with $n_u$.

\begin{figure}
\begin{center}
\includegraphics[width=8cm]{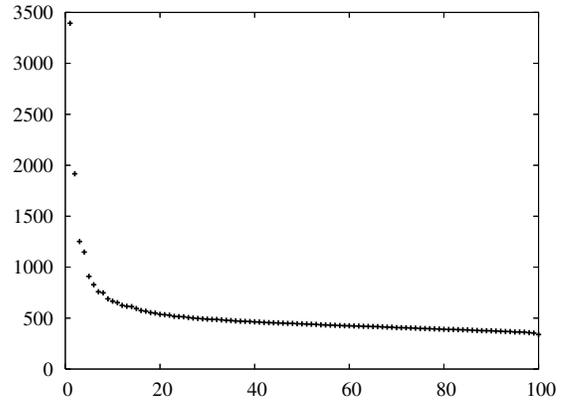}
\caption{{\small Distribution of the singular values of the complete sub matrix in the Jester joke data set. }} \label{fig:jesterjoke}
\end{center}
\end{figure}

Looking at a complete matrix where all the entries are known
can bring some insight into the structure of real data matrices. 
With Jester joke data set, we deleted all users containing 
missing entries, and generated a complete matrix $M$ with 
$14,116$ users and $100$ jokes. The distribution of 
the singular values of $M$ is shown in Figure \ref{fig:jesterjoke}. 
We must point out that this rating matrix is not low-rank or even approximately low-rank, 
although it is common to make such assumptions. This is one of  the 
difficulties in dealing with real data. The other aspect is that 
the samples are not drawn uniformly at random as commonly assumed in \cite{CandesTaoMatrix,KOM09}.

Numerical simulation results on the Movielens data set 
is also shown in the last row of the above table. 
The data set contains $100,000$ ratings for $1,682$ movies from $942$ users.\footnote{The dataset is available at {http://www.grouplens.org/node/73}} We use 
$80,000$ randomly chosen ratings to estimate the $20,000$ ratings in the test set, 
which is called $u1.base$ and $u1.test$, respectively, in the movielens data set. 
In the last row of the above table, we compare the resulting NMAE using \incoptspace, {\sc FPCA} and {\sc ADMiRA}.

%
%

%
%
\bibliographystyle{IEEEtran}

\bibliography{MatrixCompletion}

\end{document}